\documentclass{article}

     \usepackage[final]{neurips_2019}

\usepackage[utf8]{inputenc} 
\usepackage[T1]{fontenc}    
\usepackage{hyperref}       
\usepackage{url}            
\usepackage{booktabs}       
\usepackage{amssymb}
\usepackage{amsfonts}       
\usepackage{nicefrac}       
\usepackage{microtype}      
\usepackage{graphicx}
\usepackage{caption}
\usepackage{xcolor}
\usepackage{longtable}
\usepackage{enumitem}
\hypersetup{
	colorlinks,
	linkcolor={red!50!black},
	citecolor={red!60!black},
	urlcolor={blue!80!black}
}

\setcitestyle{authoryear,square}

\title{Demystifying Deep Neural Networks Through Interpretation: A Survey}
\author{Giang Dao \& Minwoo Lee\\
Department of Computer Science \\
University of North Carolina at Charlotte \\
  \texttt{gdao@uncc.edu, minwoo.lee@uncc.edu} \\}

\begin{document}

\maketitle

\begin{abstract}
  Modern deep learning algorithms tend to optimize an objective metric, such as minimize a cross entropy loss on a training dataset, to be able to learn.
  The problem is that the single metric is an incomplete description of the real world tasks. 
  The single metric cannot explain why the algorithm learn. 
  When an erroneous happens, the lack of interpretability causes a hardness of understanding and fixing the error. 
  Recently, there are works done to tackle the problem of interpretability to provide insights into neural networks behavior and thought process.
  The works are important to identify potential bias and to ensure algorithm fairness as well as expected performance.
\end{abstract}

\section{Introduction}\label{intro}
Deep neural networks have shown a broad range of success in multiple domains including image recognition tasks, natural language tasks, recommendation systems, security, and data science \cite{pouyanfar2018survey}.
Despite the success, there is a general mistrust about the system results.
Neural network prediction can be unreliable and contain biases \cite{geman1992neural}.
Deep neural networks are easy to be fooled to output wrong predictions in image classification task \cite{nguyen2015deep}.
Not only in the image recognition task, adversarial attack can be applied in natural language processing tasks \cite{jia2017adversarial}.
The problem becomes worse in security applications to secure against trojan attacks \cite{liu2017trojaning}.
Even though there have been discrimination methods developed to defend such adversarial attacks \cite{madry2017towards,carlini2017adversarial}, the unintuitive errors, which cannot fool human perception, still remain as a big problem in neural networks.
The need for demystifying neural networks has arisen to understand the neural network's unexpected behavior.

With the demand for understanding neural networks, some existing deployed systems are required to be interpretable by regulations.
The European Union has adopted the General Data Protection Regulation (GDPR) which became law in May 2018.
The GDPR stipulated ``a right of interpretability" in the clauses on automated decision-making. 
The inequality or bias, the safety of human users, industrial liability, and ethics are endangered without establishing trustworthiness based on interpretation (thus understanding) of the systems.
Therefore, the demand for interpretability created a new line of research to understand {\em why} a neural network makes a decision.
Reflecting on the needs, the number of neural networks interpretability research has been growing fast since AlexNet \cite{krizhevsky2012imagenet} came out in 2012\footnote{Google Scholar found about 18,500 results of `neural networks interpretability' from 2012 to 2020 (accessed in Feb. 17, 2020).}.


In this survey, we review existing study to interpret neural networks to help human understand what a neural network has learned and why a decision is made.
For this, we define interpretability, restate the significance, and compile them with a high-level categorization in Section~\ref{background}.
We review the interpretation methods in each category in Section~\ref{approaches}.
In Section~\ref{evaluation}, we highlight different ways to evaluate a interpretable neural network framework.
We discuss new challenges and conclude in Section~\ref{challenges}, draw conclusion in Section~\ref{conclusion}, and propose the future directions for the field in Section~\ref{future}.

\section{Definition \& Importance of Neural Network Interpretability}\label{background}
Interpretation is defined as {\em the action of explaining the meaning of something}\footnote{Accessed from Google dictionary in Feb. 17, 2020.}.
In the context of this paper, we slightly modify the definition of interpretation as {\em the action of explaining what the neural networks have learned in understandable terms to human} that anyone without deep knowledge in neural networks can understand why the neural networks make a decision.
The understandable terms are tied to knowledge, cognition, and bias of humans.
The interpretable system needs to provide information in a simple and meaningful manner.

Why is it important to understand or interpret a neural network model when it is performing well on a test dataset? 
Most of the time we don't certainly know if the dataset is generalized or covering all possibilities.
For example, self-driving car technology needs to learn a lot of accident cases to be able to generalize and perform well in the real world situation, but there can be infinite possibilities of cases that are impossible to fully collect or synthesize.
A correct prediction should be derived from a proper understanding of the original problem.
Therefore, we need to explore and understand why a neural network model makes certain decisions.
Knowing `why' helps us learn about the problem, the data, and the reasons why the model might succeed or fail.

Doshi and Kim \cite{doshi2017towards} provided reasons that drive the demand for interpretability:
\begin{enumerate}
    \item There is a big wave of change from qualitative to quantitative and toward deep neural networks with the increasing amount of data.
    In order to gain {\em scientific understanding}, we need to make the model as the source of knowledge instead of the data.
    
    \item Deploying neural networks model for automation has been increasing in real world practices.
    Therefore, monitoring the {\em safety} of the model is necessary to ensure the model operates without harming the environment.
    
    \item Despite the complexity of neural networks, encoding fairness into neural networks might be too abstract.
    Microsoft has announced the bias and discrimination problem of facial recognition\footnote{https://blogs.microsoft.com/on-the-issues/2018/12/06/facial-recognition-its-time-for-action/}. Ensuring the model {\em ethics} can increase trust from users.
    
    \item The neural networks may optimize an {\em incomplete objective}. Most of the deep neural networks minimize cross-entropy loss for classification task. However, the cross-entropy loss is known to be vulnerable to adversarial attacks \cite{nar2019cross}.
\end{enumerate}

\section{Related Work}

Some previous papers have surveyed on interpreting machine learning in different domains.
The trend in interpretable artificial intelligence in human-computer interface research by reviewing a large number of publication records \cite{abdul2018trends}.
Reviewing a large number of articles, the authors emphasized the lack of methods being applied to interpretability and encouraged a broader interpretability methods to current research.
The interpretation of a black box model has been surveyed \cite{guidotti2018survey}.
The authors divided the interpretable methods based on the types of problems: interpreting a black box model, interpreting black box outcomes, inspecting a black box, and designing a transparent box model.
The authors 
acknowledge that some approaches have attempted to tackle interpretability problems but
some important scientific questions still remain unanswered.

From analyzing the related works, we recognize that researchers have been focusing on interpreting deep neural network model in the modern works because deep neural network uses a lot of parameters and operations to derive a prediction with a low error rate.
For example, ResNet \cite{he2016deep} holds around 50 million parameters and performs around 100 billion operations to classify an image \cite{canziani2016analysis}.
This complex system makes the neural network difficult to interpret.
Therefore, interpretation of neural networks becomes an exciting area of research.
With the challenge in interpretibility of neural networks, we focus on surveying methods of how to interpret a neural network model to fully understand why the neural network makes its decision.
We go deeper and highlight different methods with their advantages and disadvantages in the sub-fields of neural networks interpretation in the next sections. 
We also provide an overview of how we can evaluate an interpretation system and propose new challenges in the interpretation field.

\section{Approaches}\label{approaches}
\begin{figure}[ht]
    \centering
    \includegraphics[width=0.95\textwidth]{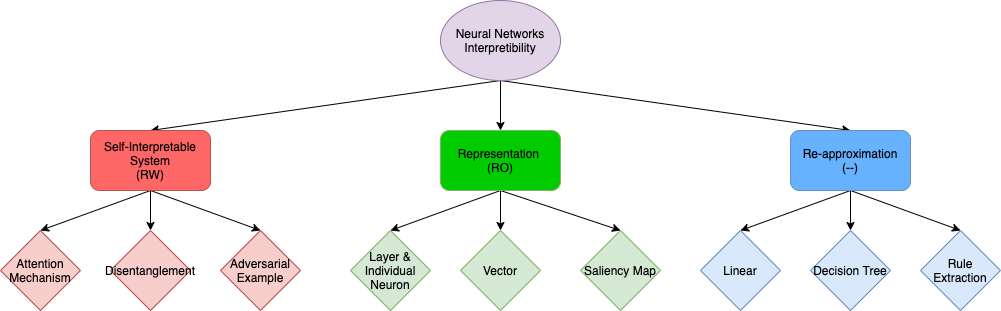}
    \caption{Splitting neural networks interpretability approaches into sub-categories and its methods of interpretation. We denote the required the accessibility to the model for interpretation: RW means read/write, RO means read-only, and -- means no access requirement.}
    \label{fig:nn_interpret}
\end{figure}
\textbf{Fig.~\ref{fig:nn_interpret}} depicts a high-level view of interpretability research in neural networks.
There exists three main approaches to interpret neural networks. 
We categorize these three main branches by how much accessibility and permission a method needs to have to interpret a neural network model: requiring full access and modification (\textit{Self-interpretable System}), requiring full access without modification (\textit{Representation Analysis}), or requiring no access or modification privilege (\textit{Re-approximation}) as follows:  
\begin{enumerate}
    \item {\em Self-Interpretable System} is a method that designs a neural network in a way that it can somewhat explain its decision. This approach requires to fully access the model to be able to modify and architect the neural network.
    \item {\em Representation Analysis} is an approach to understand individual sub-system inside the neural network by simply observing the weights and gradient updates. As it is not necessary to modify the neural network model, only full read access is enough for methods in this category.
    \item {\em Re-approximation} uses genuinely interpretable models to understand the neural networks. This approach does not read or modify the model to understand it. It simply monitors input and output of the model and re-approximates for interpretation.
\end{enumerate}

We compiled all approaches and methods that we reviewed with advantages and disadvantages in Table~\ref{tab:full}.

We split the interpretibility system into three main branches because of the user accessibility to the neural networks.
For example, a neural network's creator can use all of the three branches to explain their model which they can modify the model to have better understanding.
Users, who download models online for their application, cannot modify the model but can access the internal to understand the model's weights.
Application programing interference (API) users, who call a neural networks API to get a result, can only understand the model by approximating it. 

\begin{table*}[ht]
\centering
\captionsetup{justification=centering}
\begin{tabular}{|p{2.1cm}|p{2.5cm}|p{3.7cm}|p{3.7cm}|}
    \hline
    \textbf{Approach} & \textbf{Method} & \textbf{Advantage} & \textbf{Disadvantage} \\
    \hline
    
    Self-Interpretable System (RW) & Attention Mechanism &
    \begin{tabular}[t]{@{\textbullet~}p{3.4cm}@{}}
         Easy to interpret which input information is relevant to output
    \end{tabular}  &
    \begin{tabular}[t]{@{\textbullet~}p{3.4cm}@{}}
         Create more parameters for training \\
         Model design is required
    \end{tabular}
    \\\cline{2-4}
    &Disentanglement &\begin{tabular}[t]{@{\textbullet~}p{3.4cm}@{}}
         Easy to understand from low dimension
    \end{tabular}  &
    \begin{tabular}[t]{@{\textbullet~}p{3.4cm}@{}}
         Limited knowledge in feature roles without examining
    \end{tabular}
    \\\cline{2-4}
    &Adversarial Example &\begin{tabular}[t]{@{\textbullet~}p{3.4cm}@{}}
         Understand neural network's vulnerability
    \end{tabular}  &
    \begin{tabular}[t]{@{\textbullet~}p{3.4cm}@{}}
         Hard to understand the meaning of the added noise
    \end{tabular}
    \\\hline
    
    Representation Analysis (RO) & Layers \& Individual Neurons Analysis&
    \begin{tabular}[t]{@{\textbullet~}p{3.4cm}@{}}
         Visualizing what features have been learned
    \end{tabular}  &
    \begin{tabular}[t]{@{\textbullet~}p{3.4cm}@{}}
         Too many visualizations for analyzing one sample
    \end{tabular}
    \\\cline{2-4}
    &Vectors Analysis &\begin{tabular}[t]{@{\textbullet~}p{3.4cm}@{}}
         Easy to understand sample distribution from visualization
    \end{tabular}  &
    \begin{tabular}[t]{@{\textbullet~}p{3.4cm}@{}}
         Current methods are not good enough\\
         Might not explain why and how each data is clustered
    \end{tabular}
    \\\cline{2-4}
    &Saliency Map &\begin{tabular}[t]{@{\textbullet~}p{3.4cm}@{}}
         Highlight important input information
    \end{tabular}  &
    \begin{tabular}[t]{@{\textbullet~}p{3.4cm}@{}}
         Noisy interpretation of input features
    \end{tabular}
    \\\hline
    
    Re-approximation (--) & Linear Approximation &
    \begin{tabular}[t]{@{\textbullet~}p{3.4cm}@{}}
         Simple to implement
    \end{tabular}  &
    \begin{tabular}[t]{@{\textbullet~}p{3.4cm}@{}}
         Slow to train for a single sample $\rightarrow$ hard to scale \\
         Lower performance to neural network
    \end{tabular}
    \\\cline{2-4}
    &Decision Tree &\begin{tabular}[t]{@{\textbullet~}p{3.4cm}@{}}
         Easy to follow tree to get answer and understand process
    \end{tabular}  &
    \begin{tabular}[t]{@{\textbullet~}p{3.4cm}@{}}
         Complex tree structure with deep networks $\rightarrow$ hard to scale
    \end{tabular}
    \\\cline{2-4}
    &Rules Extraction &\begin{tabular}[t]{@{\textbullet~}p{3.4cm}@{}}
         Straight forward to analyze a sample
    \end{tabular}  &
    \begin{tabular}[t]{@{\textbullet~}p{3.4cm}@{}}
         Complex rules are hard to keep track $\rightarrow$ hard to scale
    \end{tabular}
    \\\hline
\end{tabular}
\caption{Full approaches and methods with their advantages and disadvantages.}
\label{tab:full}
\end{table*}

We summarize the splitted approaches and the methods with each own advantage and disadvantages in \ref{tab:full}.

\subsection{Self-Interpretable System} \label{approaches:self_system}
Several efforts have been taken to design a neural network model that is able to interpret its decisions after well-trained.
There are three main methods to design an interpretable neural networks model: {\em attention mechanism}, {\em disentanglement learning}, and {\em adversarial examples}.
An output of a specifically designed layer in the self-interpretable system can be easily understood because it is represented as a probability distribution in attention mechanism, vector space in disentanglement learning, and sample representation in adversarial examples.

\subsubsection{Attention Mechanism}\label{sub:attention}
    \begin{figure}[htbp]
        \centering
        \includegraphics[width=0.6\textwidth]{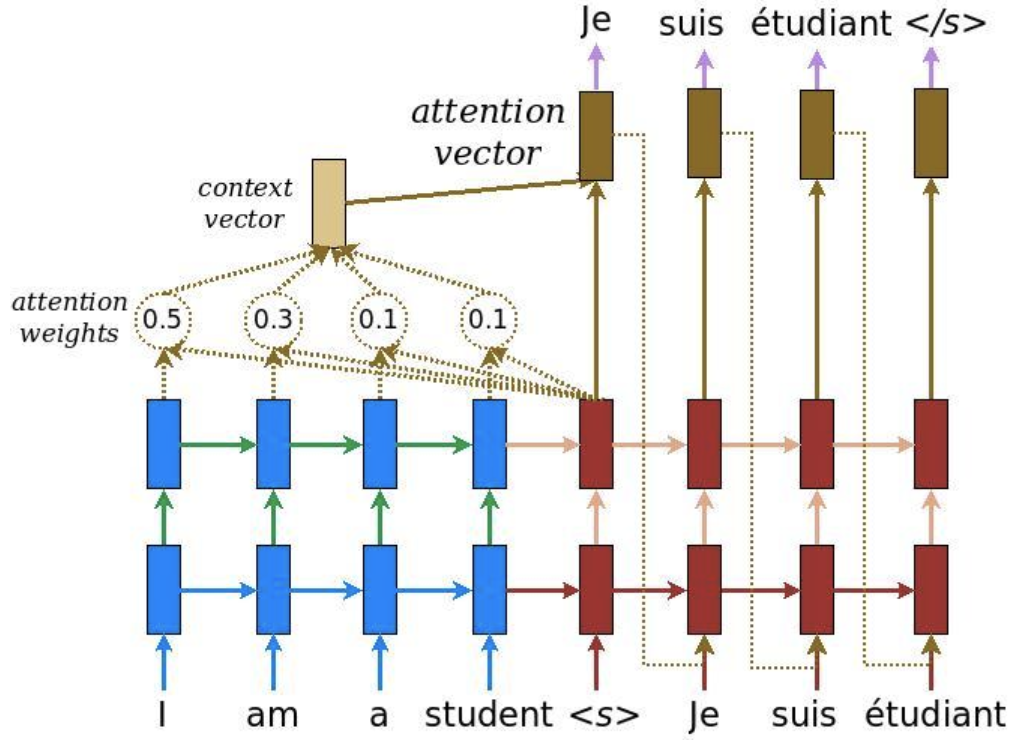}
        \caption{An example of translating from English to French showing the attention weights of the word ``Je" in French has highest correlation probability with the word ``I" in English using soft-attention from \cite{luong2015effective} method.}
        \label{fig:attention_example}
    \end{figure}
\begin{figure}[htbp]
        \centering
        \includegraphics[width=0.9\textwidth]{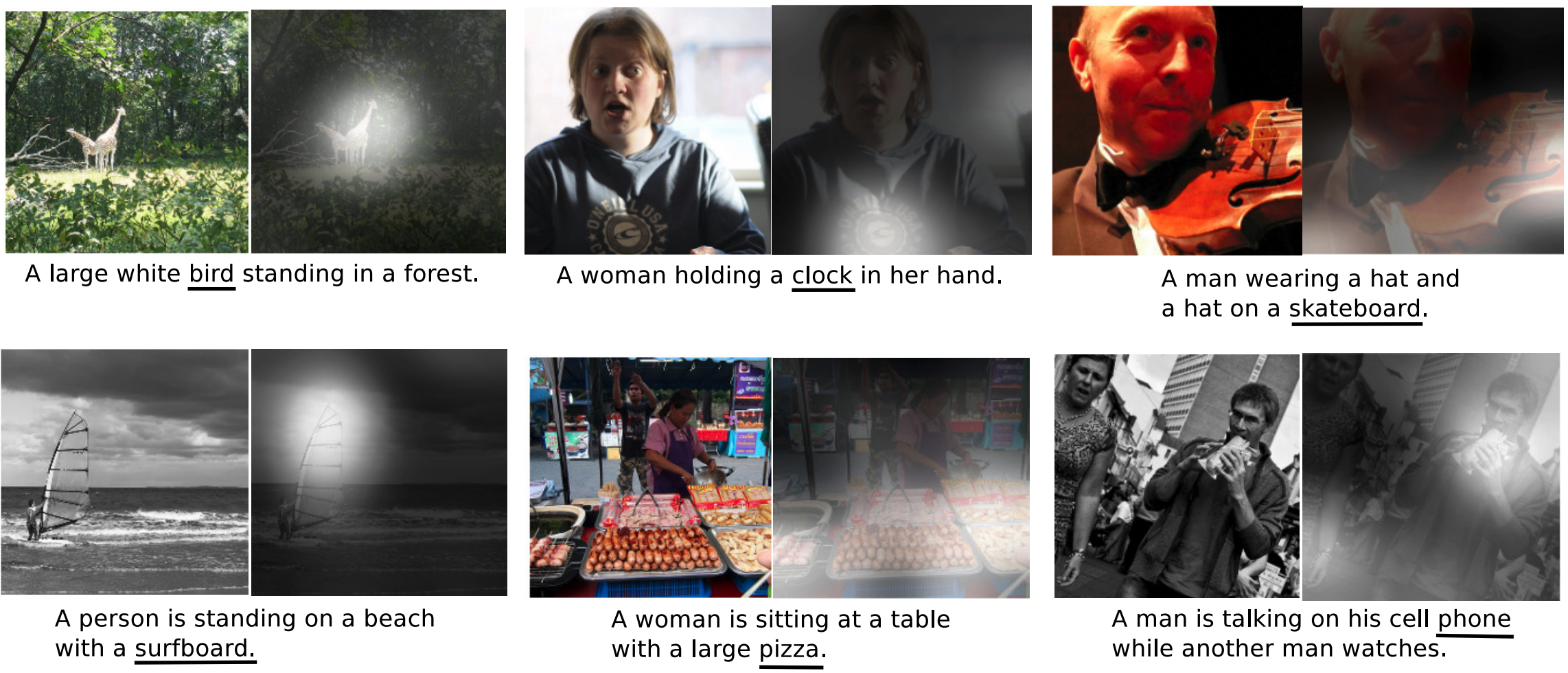}
        \caption{Visual examples interpreting why image captioning produces error by looking at the attention region proposed by \cite{xu2015show}.}
        \label{fig:show_tell_sample}
    \end{figure}

    Attention mechanism attempts to understand the relationship between information.
    Attention in deep learning is a vector of importance weights which shows how an input element correlates to target output. 
    Attention weights can be formulated as a probability distribution of correlation between a target with other sources.
    A higher probability results from a higher correlation between a target and a source.
    There are two types of attention mechanisms: hard-attention and soft-attention.
    Hard-attention strictly enforce attention weights to either $0$ for non-correlated or $1$ for correlated (Bernoulli distributions).
    Soft-attention represents attention weights with more flexible probability distributions. 
    With the flexibility, soft-attention recently dominates over hard-attention in most of the applications.
    An example of computing soft-attention weights is using softmax function to compute the correlation between a target with other sources:
        $$\alpha_{ts} = \frac{exp (score(h_t, \bar{h}_s))}{\sum_{s'=1}^{S} exp (score(h_t, \bar{h}_s))} .$$
    Attention mechanism has achieved remarkable success in natural language translation with different score functions as well as other optimization tricks \cite{graves2014neural,bahdanau2014neural,luong2015effective,canziani2016analysis}.
    A TensorFlow tutorial\footnote{https://www.tensorflow.org/tutorials/text/nmt\_with\_attention} shows an example of attention mechanism in a machine translation task in Fig.~\ref{fig:attention_example}.
    Not only showing the capability of self-interpretability in natural language processing tasks, attention mechanisms can also be designed to interpret neural network decision by looking at the attention pixels in different tasks: image classification \cite{xiao2015application,wang2017residual}, image segmentation \cite{chen2016attention}, and image captioning \cite{xu2015show,lu2016hierarchical,lu2017knowing,anderson2018bottom}.
    The neural networks error prediction can be interpreted by attention mechanism shown in Fig.~\ref{fig:show_tell_sample}.
    
    Even though attention units reveal interpretable information, they are hardly evaluated because of the robustness in the comparison process.
    Therefore, Das et al. \cite{das2017human} has created human attention datasets to compare the attention between neural networks and humans to see if they look at the same regions when making a decision.
    To enforce the neural networks to look at the same region as human and to have similar human behavior, a method to train attention mechanisms explicitly through supervised learning with the attention datasets by constraining the machine attention to be similar to human attention in the loss function was proposed \cite{ross2017right}.

    Despite the advantage of easy to interpret which input information is highly correlated to a target output, the attention mechanism caries two disadvantages.
    One is creating more parameters for training with more complex computation graph.
    The second disadvantage is that it requires the full accessibility to the model.

\subsubsection{Disentanglement Learning}
    Disentanglement learning is a method to understand a high level concepts from low level information.
    Disentanglement learning is a learning process that learns disentangled representations in lower dimensional latent vector space where each latent unit represents a meaningful and independent factor of variation. 
    For example, an image contains a black hair man will have representation of gender: male, and hair color: black encoded in the latent vector space.
    A disentangled representation can be learned explicitly from training a deep neural network.
    There are two different ways that can be considered to learn disentangled representation.
    The disentangled representation can be learned through generative adversarial networks (GAN) \cite{goodfellow2014generative} and variational autoencoder (VAE) \cite{kingma2013auto}.
    
\begin{figure}[htbp]
        \centering
        \includegraphics[width=0.9\textwidth]{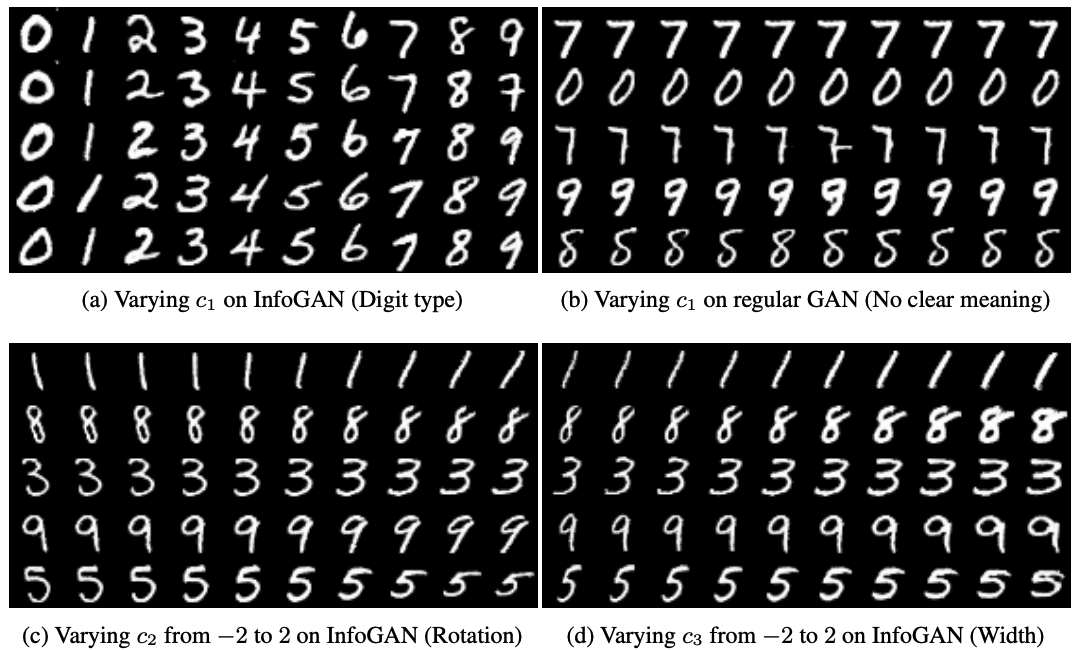}
        \caption{Interpretation result of InfoGAN \cite{chen2016infogan} by adjusting different parameters in latent dimension with different effects on the produced images. The figure shows the first latent unit corresponds for different digit type (a), the second latent unit handles the rotation of the digit (c), and the third latent unit manages the width of the digit (d). The authors also compared with the original GAN to shows the interpretibility by manipulating latent dimension (b).}
        \label{fig:infogan}
    \end{figure}

    GAN contains 2 main parts (generator and discriminator) which learns to map a vector representation into higher dimensional data. 
    The generator takes a vector representation to generate a data point. 
    The vector representation usually has lower dimension than the generated data point.
    The discriminator takes a data point and outputs true if the data is real and false if the data is generated.
    After the learning process, the vector representation usually provides high level information of the data. 
    InfoGAN \cite{chen2016infogan} is a scalable unsupervised approach to increase the disentanglement by maximizing the mutual information between subsets of latent variables and observations within the generative adversarial network.
    Auxiliary classifier GAN \cite{odena2017conditional} extends InfoGAN by controlling a latent unit with actual categorical classes. This is simply adding a controllable disentangled unit with a known independent factor.
    Fig.~\ref{fig:infogan} shows the output is varied when tuning only one latent unit of InfoGAN. 
    
    Instead of learning to map a vector representation into a data point, VAE learns to map a data point to a lower vector representation.
    VAE minimizes a loss function: 
        $$\mathcal{L}(\theta, \phi, x) = \frac{1}{L} \sum^L_{l=1} (log p_\theta(x|z^l)) - D_{KL}(q_\phi (z|x) || p_\theta(z)),$$
    has been shown as a promising direction to explicitly learn disentanglement latent units with $\beta$-VAE \cite{higgins2017beta}.
    $\beta$-VAE magnifies the KL divergence term with a factor $\beta > 1$:
        $$\mathcal{L}(\theta, \phi, x) = \frac{1}{L} \sum^L_{l=1} (log p_\theta(x|z^l)) - \beta D_{KL}(q_\phi (z|x) || p_\theta(z)),$$
    Further experiment \cite{burgess2018understanding} showed the disentangled and proposed modification of KL divergence term in the loss function to get improvement in reconstruction:
        $$ \mathcal{L}(\theta, \phi, x) = \frac{1}{L} \sum^L_{l=1} (log p_\theta(x|z^l)) - \beta |D_{KL}(q_\phi (z|x) || p_\theta(z)) - C|,$$
    with $C$ is a gradually increasing number to a large enough value to produce good reconstructions.
    The first term, $\frac{1}{L} \sum^L_{l=1} (log p_\theta(x|z^l))$, is an expected negative reconstruction error, while the second term, Kullback-Leibler divergence of approximate posterior from the prior $D_{KL}(q_\phi (z|x) || p_\theta(z))$, acts as a regularizer.
    The $\beta$ magnifies the KL divergence term to have better constrain on the prior and the posterior.
    Since KL divergence term can grow to infinity, the gradually increasing number $C$ makes the term stay numerically computable.
    
    Both GAN and VAE methods can be trained in such a way that each individual latent unit is corresponding to a specific feature.
    \cite{van2019disentangled} observed the disentangle learning leads to a better abstract reasoning.
    Graph construction (\cite{zhang2017growing}) and decision trees (see more in Section~\ref{approaches:reapprox}) are additional methods using disentangle latent dimensions.
    High-level concepts can also be represented by organizing the disentanglement with capsule networks by \cite{sabour2017dynamic}.
    Disentanglement learning is not only designed for interpretability, it recently shows huge improvement in unsuppervised learning tasks via encoding information (\cite{oord2018representation,lowe2019putting}).
    
    The disentanglement learning has an advantage of low dimensional representation (or interpretation) which is straightforward to understand. However, limited knowledge in the role of each dimension requires manual inspection for interpretation. For example, we cannot know exactly what the first latent unit is representing the digit type in InfoGAN without doing a repeated experiment.

\subsubsection{Adversarial Examples}
    Adversarial examples can be used for interpretation of neural networks bu revealing the vulnerability of the neural networks. 
    An adversarial attack is a method to deceive a neural network model.
    The main idea is to slightly perturb the input data to get a false prediction from the neural networks model, although the perturbed sample makes no different to human perception.
    Early work has been proposed \cite{szegedy2013intriguing} to find the perturbation noise by minimizing a loss function: 
        $$\mathcal{L} = loss(\hat{f} (x + \eta), l) + c \cdot |\eta|,$$
    where $\eta$ is the perturbed noise, $l$ is the desired deceived target label to deceive the neural networks, and $c$ is a constant to balance the original image and the perturbed image.
    Goodfellow et al. \cite{goodfellow2014explaining} proposed a fast gradient method to find $\eta$ by the gradient of the loss w.r.t to the input data: $\eta = \epsilon \cdot sign(\nabla_x \mathcal{L}(x, l))$.
    However, the two methods require a lot of pixels to be changed.
    Yousefzadeh and O'Leary \cite{yousefzadeh2019interpreting} reduced the number of pixels using flip points.
    It is al possible to deceive a neural network classifier with only one pixel change \cite{su2019one}.
    
    \begin{figure}[htbp]
        \centering
        \includegraphics[width=0.45\textwidth]{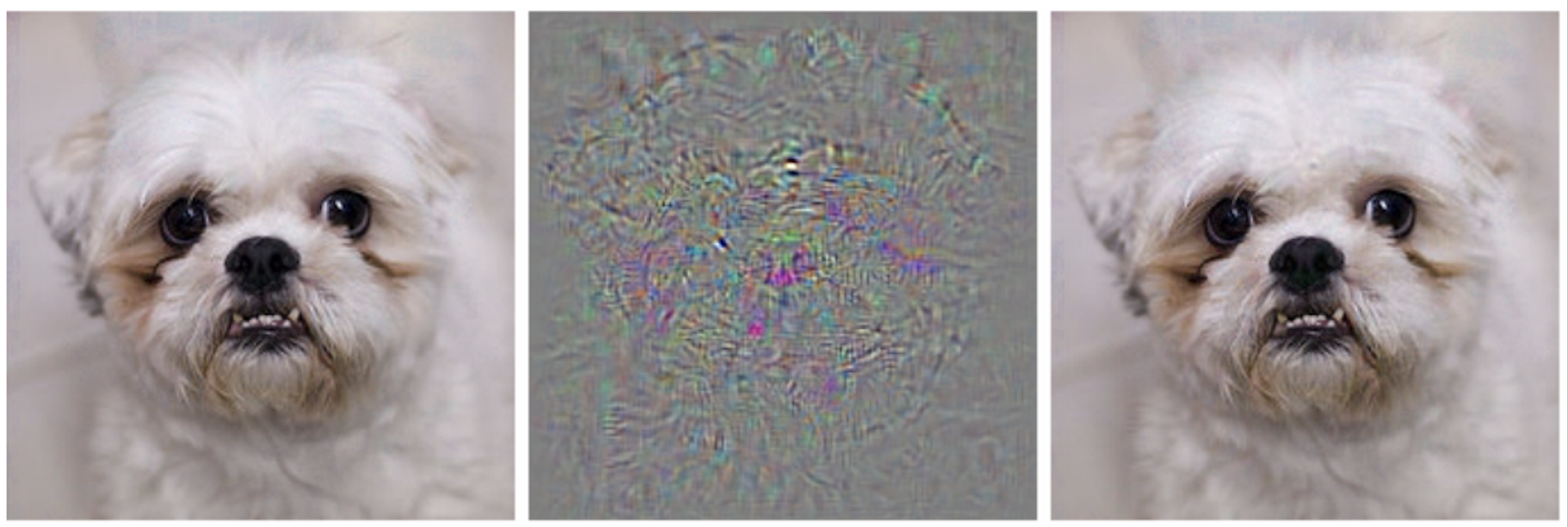}
        \includegraphics[width=0.45\textwidth]{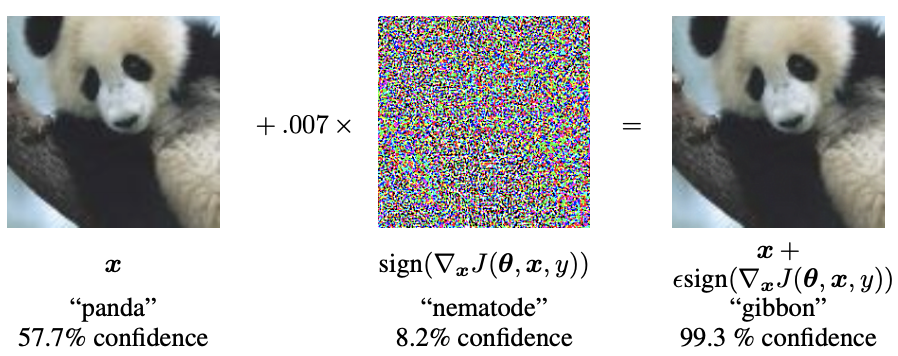}
        \caption{Left images are the original images, middle images are perturbation noise, and the right images are the perturbed images. The upper images are done by \cite{szegedy2013intriguing}, and the lower images are done by \cite{goodfellow2014explaining}. There is no different in human perception. However, The perturbed images are classified wrong by the neural networks with the desired deceived predictions.}
        \label{fig:adversarial_examples}
    \end{figure}
    
    Fig.~\ref{fig:adversarial_examples} shows how a neural networks can be deceived by changing a digital image.
    However, it is hard to intentionally modify a digital image when the image is captured by a camera without hacking into a system.
    A method to print stickers that can fool a neural networks classifier \cite{brown2017adversarial} was designed.
    Similarly, the usage of 3D printer to print a turtle but is classified as a rifle \cite{athalye2017synthesizing} has also implemented.
    
    Differently to the other neural network interpretability methods, adversarial examples focus on interpreting the vulnerability of the neural networks.
    Through different methods to generate adversarial examples, researchers observe that the neural networks are vulnerable to the adversarial examples with a small noise addition while human perception is not deceived by the adversarial examples.
    The known and discovered vulnerabilities help to enhance and to strengthen neural network decision boundaries \cite{miyato2018virtual,douzas2018effective}.
    One disadvantage of adversarial example is 
    that the meaning of the added noise is unclear to human perception and why the added noise changes the prediction of the neural network.
    

\subsection{Representation Analysis} \label{approaches:representation}
Even though there are millions of parameters and billions of computing operations, deep neural networks are internally divided by smaller subcomponents. 
The subcomponents are {\em layers \& individual neurons}, {\em vectors}, and {\em input information}.
For example, ResNet50 can be organized into 50 layers, and each layer computes between 64 to 2048 neurons.
The final layer of ResNet50 contains a vector of 2048 dimensions.
Layer, individual neuron, vector representation, and input information can interpret the decision of the neural networks.

\subsubsection{Layers \& Individual Neurons Analysis}

    \begin{figure}[htbp]
        \centering
        \includegraphics[width=0.9\textwidth] {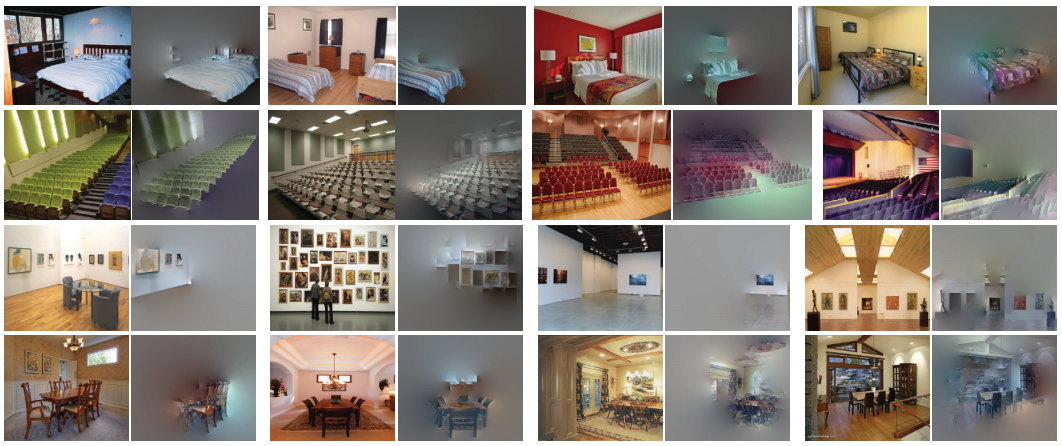}
        \caption{Pair of examples of simplified input information (right) from original input (left) image derived by \cite{zhou2014object} method.}
        \label{fig:simplified_example}
    \end{figure}
    Visualization of layer and individual neurons are helpful to understand which features have been learned.
    The information flows in neural networks can be subdivided into layers and individual neurons.
    A single individual neuron can be understood by visualizing the input patterns that maximize the neuron's response.
    With the neuron's responses visualization, researchers inpteret which information has been learned and passed through different layers and individual neurons of the neural networks.
    
    We can directly visualize each individual neurons to observe the weights.
    By visualizing and observing each layers of a small neural network, the neural network is shown to learn from simple concepts to high level concepts through each layer \cite{lee2009convolutional}.
    A neural network model first learns to detect edges, angles, contours, and corners in a different direction at the first layer, object parts at the second layer, and finally object category in the last layer.
    This sequence consistently happens during training different neural networks on different tasks.
    
\begin{figure}[htbp]
        \centering
        \includegraphics[width=0.9\textwidth] {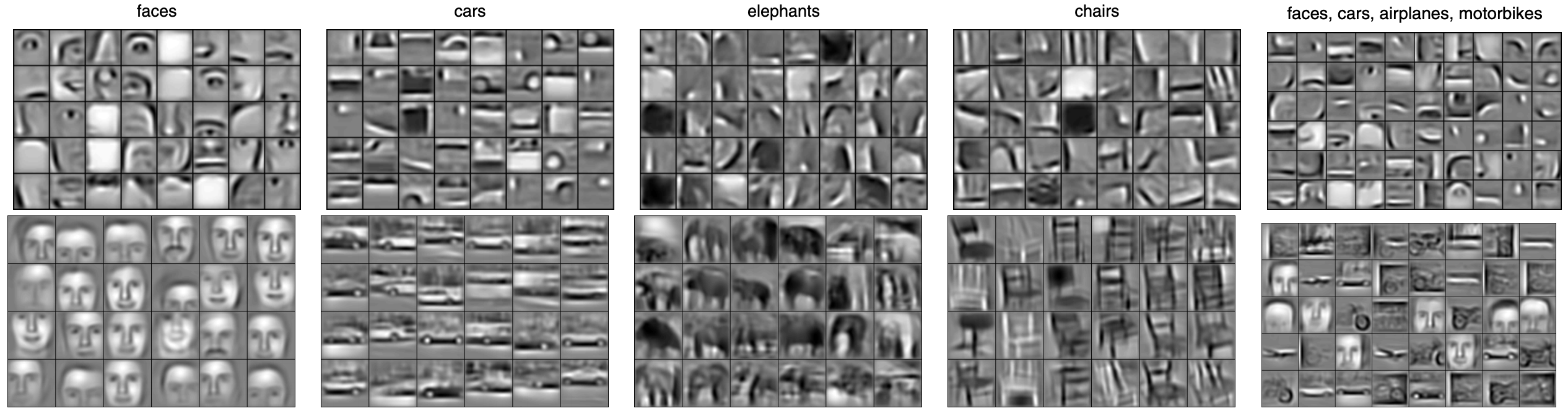}
        \caption{An illustration of what information is learned through different layers of neural networks in different tasks shown in \cite{lee2009convolutional}.}
        \label{fig:units_sample}
    \end{figure}

    Instead of visualizing neurons directly, researchers found out that the neurons' gradient can also be observed to reveal where important information parts come from.
    Gradient-based methods, which propagates through different layers and units \cite{simonyan2013deep,olah2018building}, were proposed.
    The gradient of the layers and units highlights areas in an image which discriminate a given class.
    An input can also be simplified which only reveals important information \cite{zhou2014object}. 
    Fig.~\ref{fig:simplified_example} provides examples of original image and simplified images pair.
    A method to synthesize an input that highly maximizes a desired output neuron using activation maximization \cite{nguyen2016synthesizing} by utilizing gradients.
    For example, the method can synthesize an image of lighter that the neural network classifier would maximize the probability of the lighter.
    Mordvintsev et al. \cite{mordvintsev2018differentiable} has successfully improved style transfer, which modifies a content image with a style of different image, by maximizing the activation difference of different layers.
    There is a survey of different methods for visualization of layer representations and diagnosed the representations \cite{zhang2018visual}.
    By analyzing individual neurons from a small neural network, Fig.~\ref{fig:units_sample} pointed out a strategy of how neural networks learns by visualizing all neurons.
    
    Another way to understand a single individual neuron and layers is to qualitatively validate its transferability to different tasks.
    A framework for quantifying the capacity of neural network transferability was introduced by comparing the generality versus the specificity of neurons in each layer \cite{yosinski2014transferable}.
    Network dissection method \cite{bau2017network} measures the ability of individual neurons by evaluating the alignment between individual neurons and a set of semantic concepts.
    By locating individual neurons to object, part, texture, and color concepts, network dissection can characterize the represented information from the neuron.
    
    There is a possibility of solving the same problem with smaller neural networks in roughly similar architecture.
    Large neural networks can contain a successful sub-networks without several individual neurons connected.
    Pruning individual neurons is also an exciting area of research not only in understanding neural networks \cite{frankle2018lottery}, but also improving the inference speed of the neural networks through quantization \cite{jacob2018quantization}.
    With the increase of complexity of neural network architecture to achieve state-of-the-art results, the number of layers and neurons also increases.
    More layers and neurons simply mean more human effort in validating more visualization.

\subsubsection{Vectors Analysis} \label{representation:vectors}
    Vector representations are taken before applying a linear transformation to the output from a neural network model.
    However, the vector representation most likely to have more than three dimensions which are hard to be visualized by computer.
    Vector visualization methods aim to reduce the dimension of the vector to two or three dimensions to be able to visualize by computer.
    Reducing the vector to two or three dimensions to visualize is an interesting research area.
    PCA \cite{frey1978principal} designs an orthogonal transformation method to convert a set of correlated variables into another set of linearly uncorrelated variables (called principal components). 
    The higher impact principal component has a larger variance.
    T-distribution stochastic neighbor embedding (t-SNE by Maaten and Hinton \cite{maaten2008visualizing}) performs a non-linear dimension reduction for visualization in a low dimensional space of two or three dimensions.
    t-SNE constructs low dimensional space probability distribution over pairs of high dimensional objects and minimize KL divergence with respect to the locations of the points on the map.
    
\begin{figure}[htbp]
        \centering
        \includegraphics[width=0.45\textwidth] {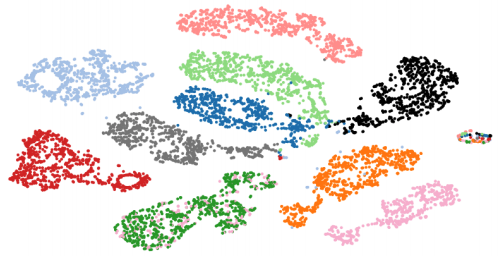}
        \includegraphics[width=0.45\textwidth] {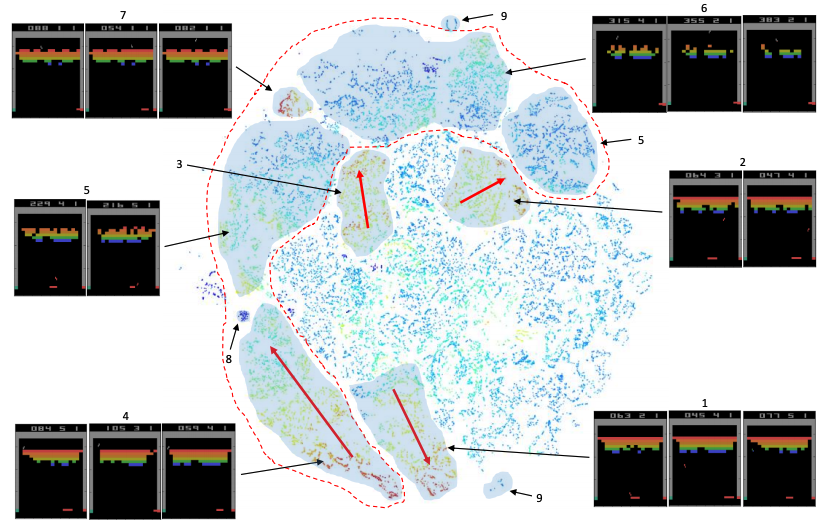}
        \caption{Examples of using t-SNE to reduce high dimension space into two dimensions to be visualizable. Left figure is showing clusters of different human voices by \cite{oord2018representation}. Right figure is different regions of action decision from a reinforcement learning agent by \cite{zahavy2016graying}.}
        \label{fig:oord_sample}
    \end{figure}

    Vector representation visualization methods are well known for helping humans understand high dimensional data.
    For example, if a neural network performs well in a classification task, the vector representations need to be clustered together if they have a similar label.
    In order to ensure the vector representations are clustered, human needs to visualize the vector and validates the assumption, especially in unsupervised learning where no label is given.
    Both of the methods reduce high dimensional space to lower dimensions (usually two or three) for an easy visualization that helps human understand and validate the neural networks.
    PCA and t-SNE are widely used by researchers to visualize high dimension information.
    As we observe Fig.~\ref{fig:oord_sample}, although the t-SNE performs reasonable well to lower the dimensions, there are areas that it does not show full separation.

\subsubsection{Saliency Map}
    Saliency map reveals significant information that affects the model decision.
    Zeiler and Furgus exemplified the saliency map by creating a map shows the influence of the input to the neural network output \cite{zeiler2014visualizing}.
    There are different techniques built upon the saliency map which showing highly activated areas or highly sensitive areas.
    The saliency method requires the direct computation of gradient from the output of the neural network with respect to the input.
    However, such derivatives are not generalized and can miss important information flowing through the networks.
    
    \begin{figure}[htbp]
        \centering
        \includegraphics[width=0.9\textwidth] {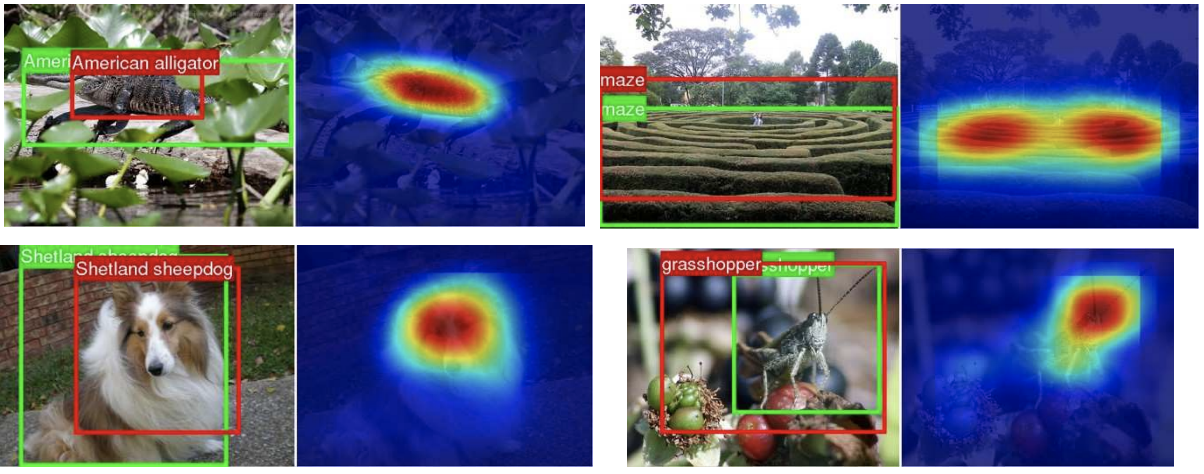}
        \caption{Pair of examples showing object detection (left) with green is ground truth and red is predicted by thresholding the salience map (right) from \cite{zhou2016learning}.}
        \label{fig:deep_features_sample}
    \end{figure}
    
\begin{figure}[htbp]
        \centering
        \includegraphics[width=0.45\textwidth] {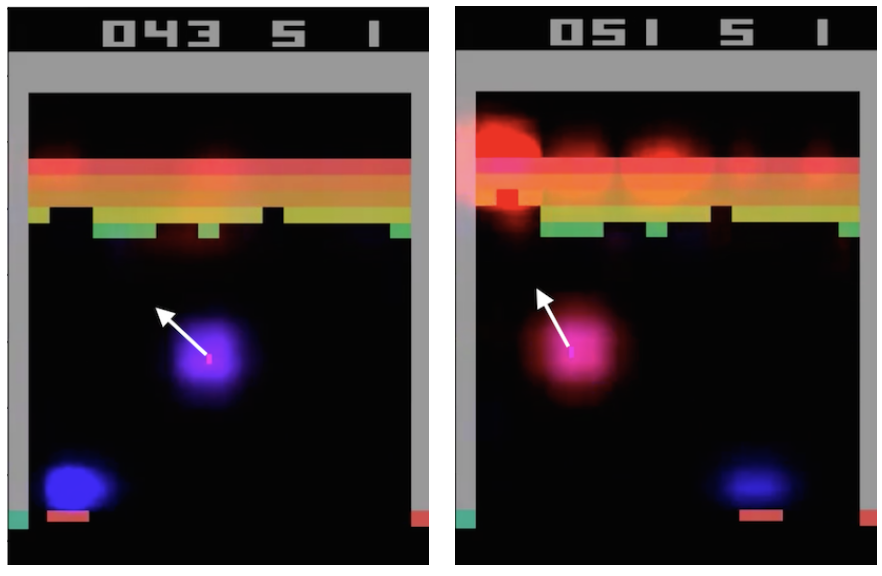}
        \caption{\cite{greydanus2017visualizing} shows how a Breakout agent learns to tunnel for high reward regions. Blue areas interpret action related regions, and red area show the areas relation with a high reward.}
        \label{fig:atari_salience_sample}
    \end{figure}

    Researchers have been working on the solution to smoothly derive the required gradient for the saliency map.
    Layer-wise relevance propagation \cite{bach2015pixel} is a method to identify contributions of a single pixel by utilizing a bag-of-words features from neural network layers.
    By simply modifying the global average pooling layer combined with class activation mapping (CAM), a good saliency map is shown \cite{zhou2016learning} comparable to an object detection method with interesting results as shown in Fig.~\ref{fig:deep_features_sample}.
    DeepLIFT \cite{shrikumar2017learning} compares the activation of each neuron with reference activations and assigns contribution scores based on the difference.
    A weighted method is used for CAM to smooth the gradient \cite{selvaraju2017grad}.
    An integrated gradient method is used to satisfy the sensitivity and implementation variance of the gradient \cite{sundararajan2017axiomatic}.
    De-noising the gradient by adding noise to perturb original input then average the saliency maps collected \cite{smilkov2017smoothgrad} also shows a better saliency map.
    An application of using saliency map to interpret why a deep reinforcement learning agent behaves \cite{greydanus2017visualizing}.
    The agent interpretable samples can be seen in Fig.~\ref{fig:atari_salience_sample} to understand the reason behind what strategy the agent has learned.
    
\subsection{Re-approximation with Interpretable Models } \label{approaches:reapprox}

By reducing the complexity of a neural network model, the networks can be interpreted efficiently. 
This has been done mainly through \textbf{re-approximation} of the neural networks with existing interpretable models.
The re-approximated model extracts the reasoning of what the neural networks have learned.
This approach works regardless of the accessibility of the neural network models, i.e., only behavioral output is enough to prepare re-approximation model for interpretation. 
There are three main methods to perform the re-approximation: {\em linear approximation}, {\em decision tree}, and {\em rules extraction}.

\subsubsection{Linear Approximation}
    A linear model can be he most simplified model that can provide interpretation of the observable outcomes.
    Linear model uses a set of weights $w$ and bias $b$ to make prediction: $\hat{y} = wx + b$.
    The linearity of the relationship between features, weights, and targets makes the interpretation easy.
    We can analyze the weights of the linear model to understand how an individual input feature impacts the decision.
    
    \begin{figure}[htbp]
        \centering
        \includegraphics[width=0.9\textwidth] {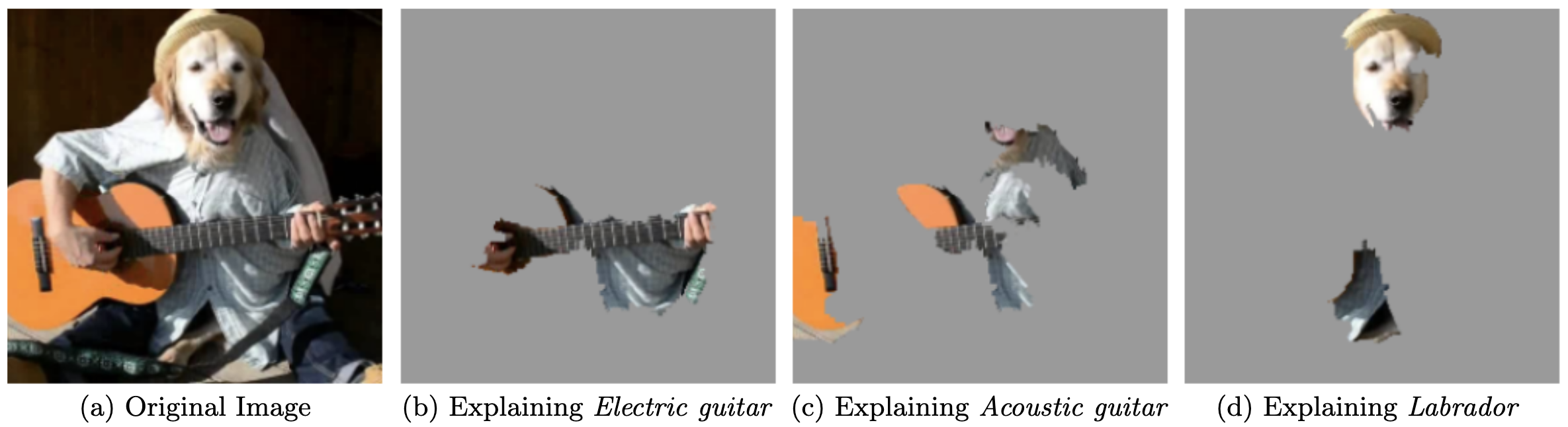}
        \caption{An example of LIME by \cite{ribeiro2016should} explains an image classification prediction from Google's Inception neural networks (\cite{szegedy2015going}) with the top 3 highest probability features: Electric guitar, acoustic guitar, and labrador.}
        \label{fig:lime_example}
    \end{figure}
    
    Local Interpretable Model Agnostic (LIME) \cite{ribeiro2016should} exemplified the linear approximation approach to classification problems.
    LIME first perturbs input data to probe the behavior of the neural networks.
    A local linear model is trained through the perturbed input and neural network output on the neighborhood information of the input.
    Fig.~\ref{fig:lime_example} shows an example of LIME identifying regions of the input that influences the neural network decision.
    
    With the simplicity in modeling, a linear approximation is by far the easiest method to implement to approximate a neural network.
    However, the linear model is hard to achieve the equivalent performance of the neural networks.
    Perturbing the neighborhood information can take a long time to train in high dimensional data.
    This makes the linear method hard to scale to the complex problems.

\subsubsection{Decision Tree}
    Linear approximation assumes input features to be independent.
    Therefore, linear approximation fails when features interact with each other to form a non-linear relationship.
    Decision trees split the data multiple times according to certain cutoff values in the data features.
    The approach results in an algorithm similar to nested if-then-else statements to compare (smaller/bigger) input features with corresponding threshold numbers.
    The interpretation is fairly simple by following the instruction from the tree root node to the leaf node.
    All the edges are connected by `AND' operation.
    
    Artifitial Neural Networks - Decision Tree (ANN-DT) \cite{schmitz1999ann} is an early work that  converts a neural network into a decision tree.
    ANN-DT applied sampling methods to expand the training data using nearest neighbors to create the decision tree.
    Sato and Tsukimoto designed Continuous Rule Extractor via Decision tree (CRED) to interpret shallow networks \cite{sato2001rule}.
    By applying RxREN \cite{augasta2012reverse} to prune unnecessary input features and C4.5 algorithm \cite{quinlan2014c4} to create a parsimonious decision tree, an extension of CRED into DeepRED \cite{zilke2016deepred} is introduced to be able apply to deep neural networks.
    The decision tree method is also applied to interpret a reinforcement learning agent's decision making \cite{bastani2018verifiable}.
    
    Although a decision tree can approximate the neural networks well to accomplish faithfulness, the constructed trees are quite large which cost time and memory to be able to scale. Furthermore, the input features of the decision tree are relatively simple that helps decision tree works. However, it is harder to approximate if the input data is in high dimensional space. Therefore, decision tree approach is hard to generalize to complex input data such as audio, images, or natural languages.

\subsubsection{Rule Extraction}
    Similar to decision trees, rule extraction methods use  nested if-then-else statements to approximate neural networks.
    While decision trees tell a user where to follow (left or right) in each node, the rule-based structures are sequences of logical predicates that are executed in order and apply if-else-then statements to make decisions.
    We can transform a decision tree to a rule-based structure and vice versa.
    Rule extraction is a well-studied approach in decision summarization from neural networks \cite{andrews1995survey}.
    There are two main approaches to extract rules from neural networks: decompositional and pedagogical approaches.
    
    Decompositional approaches mimics every individual unit behavior from neural networks by extracted rules.
    Knowledgetron (KT) method \cite{fu1994rule} sweeps through every neural unit to find different thresholds and apply if-then-else rules.
    The rules are generated based on input rather than the output of the preceding layer in a merging step.
    However, the KT method has an exponential time complexity and is not applicable to deep networks.
    The KT method was improved to achieve the polynomial time complexity \cite{tsukimoto2000extracting}.
    Fuzzy rules was also created from neural network using the decompositional approach \cite{benitez1997artificial}.
    Towell et al. \cite{towell1993extracting} proposed M-of-N rules which explain a single neural unit by clustering and ignoring insignificant units.
    Fast Extraction of Rules from Neural Networks (FERNN) \cite{setiono2000fernn} tries to identify meaningful neural units and inputs.
    Unlike other reapproximation methods, the aforementioned decompositional approaches require a full access to the information of neural network models.
    
    Pedagogical approaches are more straightforward than decompositional approaches by extracting rules directly from input and output space without sweeping through every layers and units.
    Validity interval analysis \cite{thrun1995extracting} identifies stable intervals that have the most correlation between input and output to mimic behavior of the neural networks.
    the pedagogical approach can also use sampling methods \cite{craven1996extracting,taha1999symbolic,johansson2005automatically} to extract the rules.
    
    Similar to decision trees, rule extraction methods are easy to analyze a sample. 
    However, the rule extraction methods can extract very complicated rules to explain a decision from deep neural networks.
    Therefore, rule extraction is also very hard to scale and generalize to the problems with complex input data.

\section{How to Evaluate an Interpretable System?}\label{evaluation}
\begin{table}[ht]
    \centering
    \begin{tabular}{|c|c|}
        \hline
        \textbf{Self-Interpretable} & Human evaluation \\
        \textbf{System} & Model bias \\
        \hline
        \textbf{Representation} & Performance by substitute task \\
         & Model bias \\
        \hline
        \textbf{Re-approximation} & Performance to original model \\
        & Performance by substitute task \\
        \hline
    \end{tabular}
    \caption{Evaluations for different interpretation approaches in our survey.}
    \label{tab:evaluation}
\end{table}

The three different categories of neural network interpretations have unique characteristics that are different from each other (e.g., the different level of accessibility to the networks).  
Therefore, there needs to be different evaluation criteria to explain how well the interpretation developed.
Table~\ref{tab:evaluation} shows the suggested evaluations for each interpretation approach.
In our survey, the four different evaluation metrics have appeared consistently:
\begin{enumerate}
    \item {\em Performance to original model}: This metric is mostly applied in the re-approximate method to compare the performance of the replaced model against the original neural network model.
    \item {\em Performance by substitute tasks}: Since some interpretation is not reflected by a neural network model, it requires different metrics to compare different attributes of the interpretations.
    \item {\em Model bias}: We can detect the bias of neural networks by testing the sensitivity of a specific phenomenon. If the sensitivity is not consistent across different relevant input information, the neural network is considered biased to a specific pattern.
    \item {\em Human evaluation}: Human is the most reliable evaluation metric. We can crosscheck the output of the interpretation method with human perception into the same problem. Human can also perform the previous three evaluation metrics.
\end{enumerate}

Human evaluation and model bias are frequently used evaluation criteria for {\em self-interpretable system} approaches.
Humans can double-check the result interpreted by the system to compare the interpretation with human perception.
For example, attention mechanism can be used for comparing human attention to details; latent space can be evaluated its dimension effect with human analysis; human perception can be used for validating the vulnerability of the neural networks with adversarial examples.
Since self-interpretable system is inside the neural networks, model bias evaluation can help the detection bias of the neural networks.
For example, attention mechanism fails to translate languages because of the bias (high probability) of a specific pair.

{\em Representation} can be interpreted by the produced visualization or presentation.
The methods can be evaluated by performance by a substitute task and model bias criteria.
We can check the performance by substitute task by checking layers and individual neurons with different inputs to see how neural networks model performs.
The same approach can be used for characterizing the layers and individual neurons' representation on a transfer task.
For example, we can compare the sensitivity of the saliency maps with brute force measurement.
The model bias method can be used to reveal models sensitivity to a specific phenomenon.
The layers and individual neurons visualization can benefit from the model bias to examine if the neural network is relying or ignoring a pattern.

The {\em re-approximation} method can be interpreted by analyzing the weights of a linear model, tracing the nodes of a decision tree, and reasoning the rules.
However, there is a trade-off between interpretability and performance in re-approximation method.
An approximated model of a neural network needs to balance between simplicity (for interpretation) and accuracy (for resemblance via accurate approximation).
Therefore, comparing the performance of the approximated model to the original neural network is a required evaluation criteria for {\em re-approximation} approach.
Researchers also compare the performance by substitute tasks by comparing the trade-off between different re-approximation methods.
Since the neural networks are much more complex than reapproximated methods, researchers tend to prefer approximate local behavior to be able to reduce the complexity of the neural networks.

\section{Challenges}\label{challenges}
The trade-off of interpreting neural network exists between the accuracy and robustness of a neural network and the meaningful or simpleness of interpretation.
The most accurate and robust model does not guarantee an interpretation of the network in an easy way. 
The simple and meaningful interpretation might not be easy to learn from a robust method. 
It is thus challenging when we do not have access to neural networks model to neither re-design nor extracting meaningful information from the model.
Reviewing the interpretation methods, we identify two challenges for interpreting neural networks: {\em robust interpretation} and {\em sparsity of analysis}.

\subsection{Robust Interpretation} \label{challenges:robustness}
Current approaches are too slow to produce robust interpretation in a timely manner.
Self-interpretable systems, even though the interpretation is fast on inference, still need to be trained for a longer time.
The representation systems need heavy computation in order to achieve visualization results.
Re-approximation methods take a long time for both training to approximate neural networks as well as produce interpretation.

Noisy interpretation can severely harm trust of the model. 
A neural network is trained from the data, possibly training data often cause erroneous interpretation because of errors in labeling process.
This phenomenon happens mostly with self-interpretable systems since the objective function designed to optimize the data-only, not the knowledge.
The objective function might not be well-covered to interpret the problem that makes the interpretation harder.
The representation methods can provide a lot of misleading information from layers and individual neurons, which are not related to human perceptions.
Re-approximation methods have limited performance compared to the original neural networks model, so misleading towards the poor interpretation.

\subsection{Sparsity of Analysis} \label{challenges:spar}
For each method, interpretations are made from individual samples or a lot of different visualizations.
If we scale up a problem with a large number of samples, a tremendous amount of observations and human effort are required.
The problem becomes worse if we interpret samples not from the dataset.
For example, in order to interpret the reasoning behind a neural network classifier, human needs to analyze different saliency maps from different input samples to validate the reasoning.
With that being said, researchers should concern about sparsity of analysis by reducing the number of visualizations that human needs to analyze.
The sparsity is one of the main challenge that we need to address to lessen human arduous effort in interpreting neural networks due to the large amount of data as well as computation units.
We need to have a method to recognize a meaningful smaller subset of the whole dataset to interpret.
From the meaningful subset, we also need figure out an interpretation between the relationship from different samples with different subsets.

\section{Conclusion}\label{conclusion}
Single metric to optimize in deep learning algorithm cannot reflex the complexity of the real world.
Safety and ethic are also the concerns when deploying an intelligent system.
In order to build safe and trustworthy intelligent system, we need to understand how and why a learning algorithm decides an action to help build better model understanding the real world around it.
In order to gain scientific understanding we need to transform model into a source of knowledge.

In this work, we present an overview on interpretability of deep neural networks in general.
The interpretability methods are split into three main branches according to the accessibility of users: (1) have access to model and able to modify, (2) have access to model but cannot modify, and (3) have no knowledge of the internal model.
Four methods to evaluate the interpretibility system are introduced: (1) performance to original model, (2) performance by substitute task, (3) model bias, and (4) human evaluation.
We also went deeper to explain the remaining challenges in the deep learning interpretation field.

\section{Future Direction}\label{future}
As we mentioned two different challenges in interpreting a neural networks, we want to emphasize the gap in the current interpretibility approaches: robust interpretability, sparsity of analysis.
In order to provide a fast and clear interpretation to human, the approach's robustness need to be ensured.
Reducing the amount of analysis can be a good research question since it will also reduce human evaluation time.
\cite{dao2018deep} has proposed a statistical method to identify important moments in a reinforcement learning problem.
A reinforcement learning agent might think differently than human but remains more effective, understanding the reason behind it can benefit a lot of areas with newly discovered knowledge.

The interpretability has been shown to be helpful to create better solutions to improve existing methods.
For example, MEENA chatbot \cite{adiwardana2020towards} achieved near human sensibleness and specificity understanding in natural language.
The interpretability in the self-interpretable system and representation can help validating the neural network predictions.
However, self-interpretable and representation systems require accessing and modifying neural networks.
In order to trust the interpretation, understanding the networks without accessing it is neccessary.
Therefore, we believe re-approximation with interpretable models is the most important approach needed to be improved in the future.

Another area we need to have an explanation in the learning model is reinforcement learning.
Reinforcement learning (RL) has actively used deep neural networks and has successfully applied to many areas such as 
playing video games \cite{mnih2015human}, robotics \cite{chen2017socially}, advertising \cite{zhao2018deep}, and finance \cite{deng2016deep}.
However, RL agents have not been able to give confidence to the users in the real world problems because of the lack of understanding (or interpretability).
It is hard to convince to people to use an RL agent deployed in a real environment if the unexplained or not understandable behavior are repeated. 
For instance, in AlphaGo's game 2 against the world best GO player, Lee Sedol, the agent flummoxed with the 37th move, which were not easily explainable at the moment. 
There can be a huge risk applying a non-understandable RL agent into a business model, especially where human safety or cost for failure is high.
There is a huge gap to fully understand why an RL agent decides to take an action and what an agent learns from training.

The interpretibility in RL can benefit humans to explore different strategies in solving problems.
For example, DeepMind open-sourced unverified protein structures prediction for COVID-19 from their AlphaFold system \cite{senior2020improved} in the middle of the epidemic. 
The system is confirmed to make accurate predictions with experimentally determined SARS-CoV-2 spike protein structure shared in the Protein Data Bank\footnote{https://deepmind.com/research/open-source/computational-predictions-of-protein-structures-associated-with-COVID-19 (Accessed Mar. 06, 2020).}.
Understanding why the RL system makes such prediction can benefit bioinformatics researchers further understand and improve the existing techniques in protein structures to faster create better treatment before the  epidemic happens.

\bibliography{main}

\end{document}